\documentclass[runningheads]{llncs}
\usepackage{graphicx}
\usepackage{xcolor}
\definecolor{tabG}{HTML}{004d00}
\definecolor{tabR}{HTML}{de3238}
 \usepackage{multirow}
\begin{document}

\title{Two Steps Forward and One Behind:\\ Rethinking Time Series Forecasting\\ with Deep Learning}

\titlerunning{Two Steps Forward and One Behind}

\author{Riccardo Ughi \and
Eugenio Lomurno \and
Matteo Matteucci}
% \author{Riccardo Ughi\orcidID{0000-0000-0000-0000} \and
% Eugenio Lomurno\inst{1}\orcidID{0000-0003-4007-3207} \and
% Matteo Matteucci\inst{1}\orcidID{0000-0002-8306-6739}}
% \author{Anonymous Authors}

\authorrunning{R. Ughi et al.}
% \authorrunning{Anonymous Authors}

\institute{Politecnico di Milano, Milano, Italy\footnote{This paper is supported by the FAIR (Future Artificial Intelligence Research) project, funded by the NextGenerationEU program within the PNRR-PE-AI scheme (M4C2, investment 1.3, line on Artificial Intelligence).}\\
\email{riccardo.ughi@mail.polimi.it, eugenio.lomurno@polimi.it, matteo.matteucci@polimi.it}}

\maketitle

\begin{abstract}

The Transformer is a highly successful deep learning model that has revolutionised the world of artificial neural networks, first in natural language processing and later in computer vision.
This model is based on the attention mechanism and is able to capture complex semantic relationships between a variety of patterns present in the input data.
Precisely because of these characteristics, the Transformer has recently been exploited for time series forecasting problems, assuming a natural adaptability to the domain of continuous numerical series.
Despite the acclaimed results in the literature, some works have raised doubts about the robustness and effectiveness of this approach.
In this paper, we further investigate the effectiveness of Transformer-based models applied to the domain of time series forecasting, demonstrate their limitations, and propose a set of alternative models that are better performing and significantly less complex.
In particular, we empirically show how simplifying Transformer-based forecasting models almost always leads to an improvement, reaching state of the art performance.
We also propose shallow models without the attention mechanism, which compete with the overall state of the art in long time series forecasting, and demonstrate their ability to accurately predict time series over extremely long windows.
From a methodological perspective, we show how it is always necessary to use a simple baseline to verify the effectiveness of proposed models, and finally, we conclude the paper with a reflection on recent research paths and the opportunity to follow trends and hypes even where it may not be necessary.

\end{abstract}
\section{Introduction}\label{Introduction}

Time series forecasting has often attracted the attention of researchers in fields as diverse as bioengineering, finance, climatology, and mechanics. With the advent of widespread data availability, it has become increasingly common to use computational models that can analyse long historical data series to identify patterns and use this information to make accurate predictions about future.
Early efforts in the field of time series forecasting were based on autoregressive statistical models such as ARIMA and SARIMA. These models were particularly effective for their predictive power, especially when used in conjunction with domain specific knowledge about the signals being predicted~\cite{ariyo2014stock}. 
At the same time, the scientific community began to explore the potential of artificial neural networks to further reduce estimation error via their adaptive capabilities.
Recurrent neural networks have emerged as a natural choice for forecasting due to their ability to extract valuable information from both feature and time domains~\cite{kong2017short,siami2019performance,yan2019hybrid}. Another popular deep learning approach to forecasting have been temporal convolutional neural networks. These networks adapt the successful paradigm of convolutional neural networks for image analysis to extract hierarchical patterns from temporal sequences and make accurate predictions~\cite{chen2020probabilistic,wan2019multivariate,lara2020temporal}.

A major turning point in the world of deep learning has been the introduction of the Transformer model~\cite{vaswani2017attention}. This model set a new benchmark for performance in a wide range of applications, including natural language processing~\cite{brown2020language,devlin2018bert}, computer vision~\cite{dosovitskiy2020image,liu2021swin}, and speech analysis~\cite{pham2019very,dong2018speech}. The key to the Transformer success lies in its attention mechanism which uses a sophisticated representation of the input and a large amount of training data to identify complex spatial and temporal correlations. These correlations are then used during the learning process to improve prediction quality. The Transformer model has since been widely adopted and has inspired numerous variations and extensions.
More recently, the Informer model has emerged as a leading alternative to traditional forecasting techniques~\cite{zhou2021informer}. Drawing heavily on the structure of the Transformer model, the Informer has been designed to be computationally efficient while maintaining high levels of performance. The authors believe that this model is well suited to predicting very long sequences, and, since its introduction, numerous Transformer-based techniques have been proposed and continue to set new standards for time series forecasting.

Despite the widespread adoption of these models and the interest they have generated, there are critical issues that leave room for questions and doubts.
Firstly, these models are often compared with each other without a baseline or a common reference, so there is no way of knowing whether they actually work well or badly.
Secondly, as Zeng \textit{et al.}~\cite{zeng2022transformers} points out, there are cases where embarrassingly simple models can not only compete with, but even outperform, the Transformer-based model of the moment.
For these reasons, and in order to shed some light on this extremely important area of research, in this article we examine the effectiveness of the most popular time series forecasting techniques presented in the literature as state of the art over the last three years.
First, we demonstrate the importance of comparing one's own new model at least against an extremely trivial baseline such as the Persistence model, show that there are models based on Transformer that perform worse than this model still being considered in the literature to be very good.
We then introduce two models derived from simplifying the Transformer, called respectively Sinformer and Sincoder, which are capable of outperforming current Transformer-based models deemed as the state of the art in forecasting.
Finally, we present two shallow models, the novel Sinusoidal Layered Perceptron (SLP) and a conventional Multi-Layer Perceptron (MLP), show how both are able to outperform any Transformer-based model, in some cases even by a large margin.
We also analyse the behaviour of these models in the presence of extremely long forecasting windows - which would be prohibitive for Transformer-based models due to memory constraints - and demonstrate their stability and robustness as predictive techniques.
We conclude our contribution with a reflection on recent research trends, which are often too focused on chasing the deep learning model of the moment, and sometimes pay little attention to promising alternatives being trapped in sort of evolutionary niches.

The rest of the paper is structured as follows:
\begin{itemize}
\item[$\bullet$] Section~\ref{Related Works} contains a description of the latest time series forecasting techniques and provides a background to the models that will be compared in the experiments.
\item[$\bullet$] Section~\ref{Method} describes our proposed models in details focusing on the simplification path that led us to extremely shallow networks.
\item[$\bullet$] Section~\ref{Experiments and Settings} provides information on the datasets and hyperparameters used in the experiments.
\item[$\bullet$] Section~\ref{Results and Discussion} describes in details the experiments conducted, comparing and commenting on the results of the proposed models against those selected from the literature. Reflections on current trends and empirically more promising alternatives are shared.
\item[$\bullet$] Section~\ref{Conclusion} concludes the article by summarising what has been discussed and laying the foundations for future research.
\end{itemize}

\section{Related Works}\label{Related Works}

The earliest efforts to apply the Transformer architecture to time series forecasting were aimed at adapting the vanilla model to a continuous domain rather than a dictionary-generated embedding, and at reducing its spatial complexity to enable longer predictions.
Shiyang Li \textit{et al.} identified two major weaknesses in the canonical Transformer architecture: first, the point-wise dot product self-attention is insensitive to local context, which can make the model vulnerable to anomalies in time series; second, the spatial complexity of the canonical Transformer grows quadratically with sequence length L, making it unfeasible to directly model long time series. To address these issues, the authors proposed a convolutional self-attention mechanism that generates queries and keys with causal convolution to better incorporate local context into the attention mechanism. They also proposed a LogSparse Transformer with a memory cost of only O(L(log L)$^2$), where L is the length of the sequence, which also improves prediction accuracy for time series with fine granularity and strong long-term dependencies under a constrained memory budget~\cite{li2019enhancing}.

In the same year, Kaiser and Levskaya proposed two techniques to improve the efficiency of Transformers. The first technique replaced point product attention with locality-sensitive hashing, reducing its complexity to O(L(log L)). The second technique used reversible residual layers allowing activations to be stored only once during training instead of N times, where N is the number of layers. The resulting model, called Reformer, showed comparable performance to standard Transformer models, while being significantly more memory efficient and faster for long sequences~\cite{kitaev2020reformer}.

Haoyi Zhou \textit{et al.} proposed an efficient Transformer-based model for forecasting long time series called Informer. The model has three key features: first, a ProbSparse self-attention mechanism that achieves O(L(log L)) time complexity and memory usage while maintaining comparable performance on sequence dependency alignment; second, a self-attention distillation that highlights dominant attention by halving the cascading layer input and efficiently handles extremely long input sequences; and third, a generative style decoder that predicts long time series sequences in a single forward operation rather than step-by-step, significantly improving the inference speed of long sequence prediction. Extensive experiments on four large-scale datasets showed that Informer significantly outperformed existing methods and provided a new solution to the long-sequence time series forecasting problem~\cite{zhou2021informer}.
Haixu Wu \textit{et al.} subsequently proposed a novel decomposition-based architecture with an auto-correlation mechanism called Autoformer. This model departs from the preprocessing convention for series decomposition and instead incorporates it as a fundamental inner block of deep models, providing Autoformer with progressive decomposition capabilities for complex time series. Drawing on stochastic process theory, the authors developed an auto-correlation mechanism based on series periodicity that detects dependencies and aggregates representations at the sub-series level. This auto-correlation block outperformed the self-attention one in terms of both efficiency and accuracy~\cite{wu2021autoformer}.

Tian Zhou \textit{et al.} recently introduced a novel method that combines the Transformer architecture with the seasonal trend decomposition technique. The decomposition method captures the global profile of the time series, while the Transformers capture more detailed structures. To further improve the performance of Transformers for long-term forecasting, the authors exploited the fact that most time series tend to have a Fourier transform decomposable representation that is sparse, but highly informative. The resulting method, known as the Frequency Enhanced Decomposed Transformer or FEDformer, is more efficient than standard Transformers with linear complexity with respect to sequence length~\cite{zhou2022fedformer}.
In recent months, Yunhao Zhang \textit{et al.} published a novel method that combines the Transformer architecture with the seasonal trend decomposition method from multivariate time series forecasting. The model, called Crossformer, exploits cross-dimensional dependency and embeds the input into a 2D vector array through Dimension-Segment-Wise (DSW) embedding to preserve time and dimensional information. The Two-Stage Attention (TSA) layer is then used to efficiently capture both cross-time and cross-dimension dependencies. By utilising the DSW embedding and the TSA layer, Crossformer establishes a Hierarchical Encoder-Decoder to use information at different scales for the final prediction. Extensive experimental results demonstrate the effectiveness of Crossformer over previous state-of-the-art methods~\cite{zhang2023crossformer}.

Despite the recent tendency to reward the latest Transformer-based model, recent studies suggest that there may be alternatives to this paradigm. Zeng \textit{et al.} argued that while Transformers are successful at extracting semantic correlations between elements in a long sequence, they may not be as effective at extracting temporal relations in an ordered set of continuous points. The authors claimed that permutation-invariant attention mechanisms, which are present in many of the aforementioned approaches, may result in a loss of temporal information. To validate their claim, the authors introduced a set of simple one-layer linear models into the comparison, namely Linear, NLinear, and DLinear. Experimental results on nine real-world datasets showed that these simple approaches surprisingly outperformed existing sophisticated Transformer-based models in all cases, often by a large margin. The authors hope that this surprising finding opens up new research directions for long time series forecasting, and advocate re-examining the validity of Transformer-based solutions~\cite{zeng2022transformers}.

\section{Method}\label{Method}

The purpose of this work is to advance the study of the long-term sequence prediction problem and to deepen the behavior of Transformer-based models advocating the use of much less complex alternatives. In this section, we will present in detail the models and techniques that we have developed and utilized in our experiments.

\subsection{Baseline}
The starting point for this work was to determine an appropriate baseline for a predictive model. While recurrent neural networks or Transformer-based models have been widely used in the literature and were previously considered the gold standard, they have limitations. These limitations include non-deterministic learning, dependence on the choice of input window and other hyperparameters, and thus the inability to be considered a robust and replicable baseline. To address these limitations, this paper proposes the use of a non parametric model, known as the \textbf{Persistence} model, as reference baseline for comparison in any forecasting problem. This model asserts that future predictions of length L will be identical to their previous L samples. Experiments in Section~\ref{Results and Discussion} will demonstrate how even this approach can compete with, and sometimes outperform, state of the art models recently presented at prestigious venues, highlighting the need for a more precise evaluation in time series forecasting.

% The starting point of this work was to determine a suitable baseline for a predictive model in this context. While recurrent neural networks or Transformer-based models have often been used in the literature and were previously considered the gold standard, they present limitations. These limitations include non-deterministic learning, dependence on the choice of input window and other hyperparameters, and thus the unfeasibility to be considered a robust and replicable baseline. To address these limitations, this paper proposes the use of the aparametric model known as the \textbf{Persistence} model as a constant of comparison in any forecasting problem. This model asserts that future predictions of length L will be identical to their previous L points. Experiments in Section~\ref{Results and Discussion} will demonstrate how even this approach can compete with and sometimes outperform state-of-the-art models recently exposed at prestigious conferences.

\subsection{Embedding}
Having defined a robust comparison baseline, we sought to determine a suitable latent representation for temporal sequences to be used as embeddings in Transformer-based models. We assumed that the presence of periodic patterns and a sufficiently large amount of data were the only two conditions necessary to extract useful information for both Transformer-based models and shallow neural networks. In this perspective, we were inspired by the Time2Vec model presented by Kazemi \textit{et al.}~\cite{kazemi2019time2vec}. This technique is intended to replace the positional encoding used in linguistic Transformers and is characterised by two linearly learnable tensors, one with periodic activation, which receive the raw temporal sequence as input and concatenate the output as a latent representation to be fed into the predictive model. The aim of Time2Vec is to extract and isolate high-level periodic patterns.
The main drawback of this technique is the increase in spatial complexity due to the concatenation operation, which can be particularly limiting in problems with very long prediction windows. To address this issue, we used a revised version of the original architecture called Additive Time To Vector (\textbf{AddT2V}) in our experiments. This variant uses the additive operator instead of concatenation, the sine function as the periodic function, and has tensors equal in size to the prediction window to map the periodicities in the input directly onto the prediction space.
From the ablation studies we have carried out, the performance of the Time2Vec and AddT2V models appears to be equivalent.

% After defining a robust comparison criterion, we sought to determine a suitable latent representation for temporal sequences to use as embeddings. Our assumption was that the presence of periodic patterns and a sufficiently large amount of data are the only two conditions necessary for useful information to be extracted for both Transformer-based models and shallow neural networks. In this context, we drew inspiration from the Time2Vec model presented by Kazemi \textit{et al.}~\cite{kazemi2019time2vec}. This technique is thought replace positional encoding used in linguistic Transformers and is characterized by two linearly learnable tensors, one with periodic activation, that receive the raw temporal sequence as input and concatenate the output as a latent representation to be fed to the predictive model. The goal is to extract and isolate high-level periodic patterns.
% The main disadvantage of this technique is the increase in spatial complexity due to the concatenation operation, which can be particularly limiting in problems with very long prediction windows. To address this issue, we employed a revised version of the architecture called Additive Time To Vector (\textbf{AddT2V}) in our experiments. This variant uses the additive operator instead of concatenation, employs the sine function as a periodic function, and has tensors sized equal to the forecasting window to directly map periodicities present in the input into the prediction space.
% From the ablation studies conducted, the performance of the models preceded by Time2Vec and AddT2V appears to be equivalent.

\subsection{Transformer-based Models}
In contrast to the literature, where Transformer-based models typically have deep and complex architectures, the new models presented in this paper aim to simplify the original Transformer architecture and analyse its behaviour in the context of long time series prediction. The most complex model, we named \textbf{Sinformer}, is a Transformer built upon a single block in both the encoder and decoder, with both blocks equal in size to the prediction window. The embedding of this model is the previously described AddT2V. To address the problem of intrinsic permutation invariance of the self-attention operation, the output of the model is a sinusoidal function rather than a linear one. The rationale behind this choice is that in an end-to-end learning context, AddT2V aims to map periodic patterns from the input to the prediction space by taking into account permutation invariant reworking and the presence of periodic components in the model output. The second Transformer-based model is a simplification of Sinformer where the decoder is removed and a sinusoidal activation is added to the encoder output. This model is called \textbf{Sencoder}.

% In contrast to the literature, where Transformer-based models typically have deep and complex architectures, the models presented in this paper aim to simplify the original Transformer architecture and analyze its behaviors in the context of long time series prediction. The most complex model, named \textbf{Sinformer}, is a Transformer composed of a single block in both the encoder and decoder, with both blocks sized equal to the forecasting window. The embedding of this model is the previously described AddT2V. To address the problem of intrinsic permutation invariance of the self-attention operation, the output of the model is a sinusoidal function rather than a linear one. The rationale behind this choice is that in an end-to-end learning context, AddT2V aims to map periodic patterns from the input to the prediction space by taking into account permutation invariant reworking and the presence of periodic components in the model output. The second Transformer-based model is a simplification of Sinformer, with the decoder removed and sinusoidal activation added to the encoder output. This model is referred to as \textbf{Sencoder}.

\subsection{Shallow Models}
Seeking for further simplification, the logical progression involves the complete removal of all components associated with the Transformer architecture, in particular the attention operators. Consequently, we have developed a model consisting of AddT2V followed by a single dense layer with the same dimensionality as the predicted window and a sinusoidal activation function. This model has been called the Sinusoidal Layered Perceptron (\textbf{SLP}). The aim of this architectural choice is to evaluate whether a less sophisticated operator such as a simple internal linear combination can result in a reduction in prediction error compared to the Sencoder model.
As a final model, we have constructed a Multi-Layer Perceptron (\textbf{MLP}) that is completely independent from the blocks previously presented and at the same time different from the simple models proposed by Zeng \textit{et al.}~\cite{zeng2022transformers}. This MLP consists of three dense layers with ReLU activation and the same dimensionality as the predicted window. The MLP model is elastically regularised with l1=10$^{-5}$ and l2=10$^{-4}$, and the last layer has a linear activation function.

% In the pursuit of further simplification, the logical progression involves the complete removal of all components associated with the Transformer architecture, particularly the attention operations. Consequently, we have developed a model that consists of AddT2V followed by a single dense layer with the same dimensionality as the predicted window and a sinusoidal activation function. This model has been designated as the Sinusoidal Layered Perceptron (\textbf{SLP}). The aim of this architectural choice is to evaluate whether a less sophisticated operators as a simple internal linear combination results in a reduction in forecasting error with respect to the Sencoder model.
% As a final model, we have constructed a Multi-Layer Perceptron (\textbf{MLP}) that is entirely independent of the previously presented blocks and at the same time distinct from the simple models proposed by Zeng \textit{et al.}~\cite{zeng2022transformers}. This MLP comprises three dense layers with ReLU activation and the same dimensionality as the predicted window. The MLP model is elastically regularized with l1=10$^{-5}$ and l2=10$^{-4}$, and the final layer has a linear activation function.
\section{Experiments Setting}\label{Experiments and Settings}

To facilitate the interpretation of the results, this section provides a detailed description of the datasets and hyperparameters used in the experiments. In addition, the models under comparison, the evaluation criteria, and the data pre-processing methods prior to the learning phase are presented.

\subsection{Datasets}
\begin{table}[t]
    \caption{This table provides a summary of the datasets utilized in the experiments conducted within this study.}
    \centering
    \resizebox{\textwidth}{!}{%
    \begin{tabular}{|c|ccc|cc|}
        \hline
        \textbf{Dataset} & \textbf{Length} & \textbf{Features} & \textbf{Frequency} & \textbf{Batch Size} & \textbf{Training/Validation/Test}\\
        \hline
        ETTh1 & 17520 & 7 & 1h & 64 & 60/20/20\\
        ETTm1 & 70080 & 7 & 15m & 64 & 60/20/20\\
        Electricity & 26304 & 321 & 1h & 32 & 70/10/20\\
        Milan T° & 7360 & 1 & 24h & 32 & 66/17/17\\
        Venice & 330000 & 1 & 1h & 256 & 60/20/20\\
        Weather & 52696 & 21 & 10m & 16 & 70/10/20\\
        \hline
    \end{tabular}}%
    \label{tab:datasets}
\end{table}

In order to gain a detailed insight into the effectiveness of the forecasting techniques under consideration, in this research we identified and selected six datasets for their heterogeneous characteristics, which are summarised in Table~\ref{tab:datasets}.
More specifically, the selected datasets are:
\begin{itemize}

\item[$\bullet$] \textbf{ETTh1 and ETTm1}~\cite{ETT}: The Electricity Transformer Temperature (ETT) datasets contain information relevant to the long-term distribution of electricity. The data were collected over a period of two years from two different counties in China. The ETTh1 dataset, which refers to the first station, consists of 17520 hourly samples. The ETTm1 dataset, also related to the first station, consists of 70080 samples taken at 15 minute intervals. Both datasets consist of eight features and were used in all proposed experiments with a training/validation/test split of approximately 14/5/5 months. Each model was trained on these datasets with a batch size of 64. These datasets represent two of the most important benchmarks for time series forecasting due to their multivariate nature and different sampling frequencies.

\item[$\bullet$] \textbf{Electricity}~\cite{electricity}: The Electricity dataset contains several features on the electricity consumption of 321 customers from 2012 to 2014. It consists of 26304 hourly samples, each with 322 features. The data was divided into training, validation and test sets of approximately 24, 4 and 8 months respectively. Each model was trained on this dataset with a batch size of 32. This dataset was chosen to represent a medium-sized multivariate dataset with a significant number of features.

\item[$\bullet$] \textbf{Temperature of Milan (Milan T°)}~\cite{temperature}: This dataset covers the daily mean temperature history of Milan from 2001 to 2021. It is the smallest dataset analysed in this study, consisting of 7360 points sampled at 24-hour intervals. The data was divided into training, validation and test sets with durations of approximately 160, 40 and 40 months respectively. Each model was trained on this dataset with a batch size of 32. This dataset was chosen to investigate the behaviour of models dealing with a small scale univariate forecasting task.

\item[$\bullet$] \textbf{High Water of Venice (Venice)}~\cite{marea}: This dataset contains the historical series of sea level values recorded in Venice from 1983 to the present. It is the largest dataset analysed in this study, consisting of 330000 hourly samples. The data were divided into training, validation and test sets with durations of 24, 8 and 8 years respectively. Due to the large amount of data, a batch size of 256 was chosen for the experiments. This dataset was chosen for its high cardinality, which allows the analysis of the behaviour of different models in a univariate context for very long forecasts.

\item[$\bullet$] \textbf{Weather}~\cite{weather}: This dataset, which contains 21 meteorological features such as air temperature and humidity, was collected in Germany in 2020. It is a medium-sized dataset consisting of 52696 samples taken at 10-minute intervals. Following the literature, the data was divided into training, validation and test sets with proportions of 70\%, 10\% and 20\% respectively. A batch size of 16 was used for the experiments. This multivariate dataset was included due to its average length and number of features relative to the other datasets.
\end{itemize}
The experiments were designed to provide a comprehensive comparison of the models' capabilities by running multivariate predictions on each multivariate dataset and univariate predictions on all others.

\subsection{Models and Setup}

% Elenco dei modelli utilizzati
The models selected for comparison on the above datasets are grouped into three categories. The first category, consisting only in the Persistence model, represents the non parametric model used as a baseline. The second category consists of shallow models, including the SLP and MLP models proposed by us and the Linear, NLinear and DLinear models proposed by Zeng \textit{et al.}~\cite{zeng2022transformers}. The third and final category consists of Transformer-based models, including the Sencoder and Sinformer models we propose in this paper, and the FEDformer~\cite{zhou2022fedformer}, Autoformer~\cite{wu2021autoformer}, Informer~\cite{zhou2021informer}, and Crossformer~\cite{zhang2023crossformer} models.

Among the various metrics used in the literature to evaluate forecasting quality, this paper uses the Mean Absolute Error (MAE) to allow a more direct comparison, independent from the magnitude of individual errors, as opposed to quadratic metrics.
In terms of the learning details of the models, the best epoch was selected according to the validation error from a training of 50 epochs performed with the Adam optimiser and an exponentially decaying learning rate starting from 10$^{-3}$ and shrinking to 10$^{-6}$. The optimal portion of the training set and the input window were determined by tuning. Data standardisation was the only form of pre-processing used. All experiments were run on a single A6000 GPU with 48 GB of memory.
\section{Results and Discussion}\label{Results and Discussion}

\begin{table}[t]
    \caption{This table presents the results of the experiments conducted on the test sets in the form of Mean Absolute Error. The evaluations were carried out over four different forecasting windows for each dataset. Results for models not presented in this paper were obtained from their respective original papers, with the exception of the results for the Milan T° and Venice datasets, which were computed as part of this paper. Crossformer results were all recalculated with the authors code. Results that outperform the baseline are coloured \textcolor{tabG}{green}, while those that underperform it are coloured \textcolor{tabR}{red}. The best results are highlighted in \textbf{bold}, while the best results considering only the Transformer-based models are \underline{underlined}.}
    \centering
    \resizebox{\textwidth}{!}{%
    \begin{tabular}{|l|c||c|ccccc|cccccc|}
        \hline
        \multicolumn{2}{|c||}{\raisebox{-6pt}{\textbf{Model}}} & \textbf{Persistence} & \textbf{SLP} & \textbf{MLP} & \textbf{Linear} & \textbf{NLinear} & \textbf{DLinear} & \textbf{Sencoder} & \textbf{Sinformer} & \textbf{FEDformer} & \textbf{Autoformer} & \textbf{Informer} & \textbf{Crossformer}\\
        \multicolumn{1}{|c}{} & & (baseline) & (ours) & (ours) & AAAI2023 & AAAI2023 & AAAI2023 & (ours) & (ours) & ICML2022 & NeurIPS2021 & AAAI2021 & ICLR2023\\
        \hline
        \multirow{4}{*}{\rotatebox[origin=c]{90}{ETTh1}} & 96 & 0.480 & \textcolor{tabG}{0.392} & \textcolor{tabG}{\textbf{0.388}} & \textcolor{tabG}{0.397} & \textcolor{tabG}{0.394} & \textcolor{tabG}{0.399} & \textcolor{tabG}{\underline{0.390}} & \textcolor{tabG}{0.415} & \textcolor{tabG}{0.419} & \textcolor{tabG}{0.459} & \textcolor{tabR}{0.713} & \textcolor{tabG}{0.426}\\
        & 192 & 0.530 & \textcolor{tabG}{0.422} & \textcolor{tabG}{0.424} & \textcolor{tabG}{0.429} & \textcolor{tabG}{\textbf{0.415}} & \textcolor{tabG}{0.416} & \textcolor{tabG}{\underline{0.433}} & \textcolor{tabG}{0.444} & \textcolor{tabG}{0.448} & \textcolor{tabG}{0.482} & \textcolor{tabR}{0.792} & \textcolor{tabG}{0.440}\\
        & 336 & 0.571 & \textcolor{tabG}{0.456} & \textcolor{tabG}{0.464} & \textcolor{tabG}{0.476} & \textcolor{tabG}{\textbf{0.427}} & \textcolor{tabG}{0.443} & \textcolor{tabG}{0.463} & \textcolor{tabG}{\underline{0.457}} & \textcolor{tabG}{0.465} & \textcolor{tabG}{0.496} & \textcolor{tabR}{0.809} & \textcolor{tabG}{0.459}\\
        & 720 & 0.718 & \textcolor{tabG}{0.528} & \textcolor{tabG}{0.540} & \textcolor{tabG}{0.592} & \textcolor{tabG}{\textbf{0.453}} & \textcolor{tabG}{0.490} & \textcolor{tabG}{0.542} & \textcolor{tabG}{0.537} & \textcolor{tabG}{\underline{0.507}} & \textcolor{tabG}{0.512} & \textcolor{tabR}{0.865} & \textcolor{tabG}{0.519}\\
        \hline
        % \multirow{4}{*}{\rotatebox[origin=c]{90}{ETTh2}} & 96 & 0.428 & \textcolor{tabG}{0.348} & \textcolor{tabG}{0.380} & \textcolor{tabG}{0.352} & \textcolor{tabG}{\textbf{0.338}} & \textcolor{tabG}{0.353} & \textcolor{tabG}{0.361} & \underline{\textcolor{tabG}{0.349}} & \textcolor{tabG}{0.388} & \textcolor{tabG}{0.397} & \textcolor{tabR}{1.525} & NA\\
        % & 192 & 0.509 & \textcolor{tabG}{0.399} & \textcolor{tabG}{0.448} & \textcolor{tabG}{0.413} & \textcolor{tabG}{\textbf{0.381}} & \textcolor{tabG}{0.418} & \textcolor{tabG}{0.445} & \textcolor{tabR}{0.773} & \underline{\textcolor{tabG}{0.439}} & \textcolor{tabG}{0.452} & \textcolor{tabR}{1.931} & NA\\
        % & 336 & 0.560 & \textcolor{tabG}{\textbf{0.381}} & \textcolor{tabG}{0.537} & \textcolor{tabG}{0.461} & \textcolor{tabG}{0.400} & \textcolor{tabG}{0.465} & \textcolor{tabG}{0.514} & \underline{\textcolor{tabG}{0.469}} & \textcolor{tabG}{0.487} & \textcolor{tabG}{0.486} & \textcolor{tabR}{1.835} & NA\\
        % & 720 & 0.682 & \textcolor{tabG}{0.455} & \textcolor{tabG}{0.681} & \textcolor{tabG}{0.595} & \textcolor{tabG}{\textbf{0.436}} & \textcolor{tabG}{0.490} & \textcolor{tabR}{0.702} & \textcolor{tabR}{0.744} & \underline{\textcolor{tabG}{0.474}} & \textcolor{tabG}{0.511} & \textcolor{tabR}{1.625} & NA\\
        % \hline
        \multirow{4}{*}{\rotatebox[origin=c]{90}{ETTm1}} & 96 & 0.389 & \textcolor{tabG}{0.335} & \textcolor{tabG}{0.334} & \textcolor{tabG}{0.352} & \textcolor{tabG}{0.348} & \textcolor{tabG}{0.343} & \textcolor{tabG}{\textbf{\underline{0.333}}} & \textcolor{tabG}{0.345} & \textcolor{tabR}{0.419} & \textcolor{tabR}{0.475} & \textcolor{tabR}{0.571} & \textcolor{tabR}{0.449}\\
        & 192 & 0.421 & \textcolor{tabG}{\textbf{0.358}} & \textcolor{tabG}{0.366} & \textcolor{tabG}{0.369} & \textcolor{tabG}{0.375} & \textcolor{tabG}{0.365} & \textcolor{tabG}{0.389} & \textcolor{tabG}{\underline{0.382}} & \textcolor{tabR}{0.441} & \textcolor{tabR}{0.496} & \textcolor{tabR}{0.669} & \textcolor{tabG}{0.413}\\
        & 336 & 0.845 & \textcolor{tabG}{\textbf{0.380}} & \textcolor{tabG}{0.403} & \textcolor{tabG}{0.393} & \textcolor{tabG}{0.388} & \textcolor{tabG}{0.386} & \textcolor{tabG}{\underline{0.415}} & \textcolor{tabG}{0.433} & \textcolor{tabG}{0.459} & \textcolor{tabG}{0.537} & \textcolor{tabR}{0.871} & \textcolor{tabG}{0.455}\\
        & 720 & 0.851 & \textcolor{tabG}{\textbf{0.414}} & \textcolor{tabG}{0.419} & \textcolor{tabG}{0.435} & \textcolor{tabG}{0.422} & \textcolor{tabG}{0.421} & \textcolor{tabG}{\underline{0.425}} & \textcolor{tabG}{0.455} & \textcolor{tabG}{0.490} & \textcolor{tabG}{0.561} & \textcolor{tabG}{0.823} & \textcolor{tabG}{0.528}\\
        \hline
        \multirow{4}{*}{\rotatebox[origin=c]{90}{Electricity}} & 96 & 0.477 & \textcolor{tabG}{0.217} & \textcolor{tabG}{\textbf{0.210}} & \textcolor{tabG}{0.237} & \textcolor{tabG}{0.237} & \textcolor{tabG}{0.237} & \textcolor{tabG}{\underline{0.218}} & \textcolor{tabG}{0.221} & \textcolor{tabG}{0.308} & \textcolor{tabG}{0.317} & \textcolor{tabG}{0.368} & \textcolor{tabG}{0.285}\\
        & 192 & 0.372 & \textcolor{tabG}{0.242} & \textcolor{tabG}{\textbf{0.233}} & \textcolor{tabG}{0.250} & \textcolor{tabG}{0.248} & \textcolor{tabG}{0.249} & \textcolor{tabG}{\underline{0.236}} & \textcolor{tabG}{0.242} & \textcolor{tabG}{0.315} & \textcolor{tabG}{0.334} & \textcolor{tabR}{0.386} & \textcolor{tabG}{0.313}\\
        & 336 & 0.435 & \textcolor{tabG}{0.277} & \textcolor{tabG}{0.297} & \textcolor{tabG}{0.268} & \textcolor{tabG}{\textbf{0.265}} & \textcolor{tabG}{0.267} & \textcolor{tabG}{\underline{0.296}} & \textcolor{tabG}{0.311} & \textcolor{tabG}{0.329} & \textcolor{tabG}{0.338} & \textcolor{tabG}{0.394} & \textcolor{tabG}{0.353}\\
        & 720 & 0.561 & \textcolor{tabG}{0.383} & \textcolor{tabG}{0.371} & \textcolor{tabG}{0.301} & \textcolor{tabG}{\textbf{0.297}} & \textcolor{tabG}{0.301} & \textcolor{tabG}{0.363} & \textcolor{tabG}{0.397} & \textcolor{tabG}{\underline{0.355}} & \textcolor{tabG}{0.361} & \textcolor{tabG}{0.439} & \textcolor{tabG}{0.449}\\
        \hline
        % \multirow{4}{*}{\rotatebox[origin=c]{90}{Exchange}} & 96 & 0.513 & \textcolor{tabG}{0.225} & \textcolor{tabG}{0.326} & \textcolor{tabG}{0.207} & \textcolor{tabG}{0.208} & \textcolor{tabG}{\textbf{0.203}} & \textcolor{tabR}{0.916} & \textcolor{tabR}{0.603} & \textcolor{tabG}{0.278} & \textcolor{tabG}{0.323} & \textcolor{tabR}{0.752} & NA\\
        % & 192 & 0.628 & \textcolor{tabG}{0.333} & \textcolor{tabG}{0.602} & \textcolor{tabG}{0.304} & \textcolor{tabG}{0.300} & \textcolor{tabG}{\textbf{0.293}} & \textcolor{tabR}{0.76} & \textcolor{tabR}{0.632} & \textcolor{tabG}{0.380} & \textcolor{tabG}{0.369} & \textcolor{tabR}{0.895} & NA\\
        % & 336 & 0.694 & \textcolor{tabR}{0.814} & \textcolor{tabR}{0.736} & \textcolor{tabG}{0.432} & \textcolor{tabG}{0.415} & \textcolor{tabG}{\textbf{0.414}} & \textcolor{tabG}{0.686} & \textcolor{tabR}{0.750} & \textcolor{tabG}{0.500} & \textcolor{tabG}{0.524} & \textcolor{tabR}{1.036} & NA\\
        % & 720 & 1.07 & \textcolor{tabG}{\textbf{0.499}} & \textcolor{tabG}{0.796} & \textcolor{tabG}{0.750} & \textcolor{tabG}{0.780} & \textcolor{tabG}{0.601} & \textcolor{tabG}{0.808} & \textcolor{tabG}{0.860} & \textcolor{tabG}{0.841} & \textcolor{tabG}{0.941} & \textcolor{tabR}{1.31} & NA\\
        % \hline
        \multirow{4}{*}{\rotatebox[origin=c]{90}{Milan T°}} & 96 & 1.337 & \textcolor{tabG}{0.328} & \textcolor{tabG}{0.318} & \textcolor{tabG}{0.304} & \textcolor{tabG}{0.303} & \textcolor{tabG}{0.299} & \textcolor{tabG}{0.360} & \textcolor{tabG}{\textbf{\underline{0.282}}} & \textcolor{tabG}{0.503} & \textcolor{tabG}{0.722} & \textcolor{tabG}{0.305} & \textcolor{tabG}{0.283}\\
        & 192 & 1.718 & \textcolor{tabG}{0.326} & \textcolor{tabG}{0.321} & \textcolor{tabG}{0.304} & \textcolor{tabG}{0.295} & \textcolor{tabG}{0.302} & \textcolor{tabG}{0.306} & \textcolor{tabG}{0.303} & \textcolor{tabG}{0.384} & \textcolor{tabG}{0.644} & \textcolor{tabG}{\textbf{\underline{0.293}}} & \textcolor{tabG}{0.303}\\
        & 336 & 0.543 & \textcolor{tabG}{0.312} & \textcolor{tabG}{0.328} & \textcolor{tabG}{0.314} & \textcolor{tabG}{0.295} & \textcolor{tabG}{0.310} & \textcolor{tabG}{0.309} & \textcolor{tabG}{0.296} & \textcolor{tabR}{0.584} & \textcolor{tabR}{0.602} & \textcolor{tabG}{0.298} & \textcolor{tabG}{\textbf{\underline{0.286}}}\\
        & 720 & 0.496 & \textcolor{tabG}{0.382} & \textcolor{tabG}{0.356} & \textcolor{tabG}{0.332} & \textcolor{tabG}{0.295} & \textcolor{tabG}{0.328} & \textcolor{tabG}{0.380} & \textcolor{tabG}{0.363} & \textcolor{tabG}{\textbf{\underline{0.271}}} & \textcolor{tabR}{0.904} & \textcolor{tabG}{0.321} & \textcolor{tabG}{0.308}\\
        \hline
        \multirow{4}{*}{\rotatebox[origin=c]{90}{Venezia}} & 96 & 0.936 & \textcolor{tabG}{\textbf{0.223}} & \textcolor{tabG}{0.274} & \textcolor{tabG}{0.275} & \textcolor{tabG}{0.279} & \textcolor{tabG}{0.274} & \textcolor{tabG}{0.273} & \textcolor{tabG}{0.270} & \textcolor{tabG}{0.357} & \textcolor{tabG}{0.665} & \textcolor{tabG}{0.302} & \textcolor{tabG}{\underline{0.267}}\\
        & 192 & 1.190 & \textcolor{tabG}{\textbf{0.242}} & \textcolor{tabG}{0.364} & \textcolor{tabG}{0.325} & \textcolor{tabG}{0.330} & \textcolor{tabG}{0.322} & \textcolor{tabG}{0.312} & \textcolor{tabG}{0.313} & \textcolor{tabG}{0.421} & \textcolor{tabG}{0.633} & \textcolor{tabG}{0.327} & \textcolor{tabG}{\underline{0.308}}\\
        & 336 & 0.525 & \textcolor{tabG}{\textbf{0.252}} & \textcolor{tabG}{0.394} & \textcolor{tabG}{0.359} & \textcolor{tabG}{0.372} & \textcolor{tabG}{0.357} & \textcolor{tabG}{0.342} & \textcolor{tabG}{0.343} & \textcolor{tabG}{0.492} & \textcolor{tabR}{0.853} & \textcolor{tabR}{0.766} & \textcolor{tabG}{\underline{0.331}}\\
        & 720 & 0.540 & \textcolor{tabG}{\textbf{0.262}} & \textcolor{tabG}{0.413} & \textcolor{tabG}{0.396} & \textcolor{tabG}{0.410} & \textcolor{tabG}{0.398} & \textcolor{tabG}{0.360} & \textcolor{tabG}{0.360} & \textcolor{tabG}{0.503} & \textcolor{tabR}{0.814} & \textcolor{tabR}{0.749} & \textcolor{tabG}{\underline{0.355}}\\
        \hline
        \multirow{4}{*}{\rotatebox[origin=c]{90}{Weather}} & 96 & 0.889 & \textcolor{tabG}{0.321} & \textcolor{tabG}{0.486} & \textcolor{tabG}{0.236} & \textcolor{tabG}{\textbf{0.232}} & \textcolor{tabG}{0.237} & \textcolor{tabG}{0.501} & \textcolor{tabG}{0.509} & \textcolor{tabG}{0.296} & \textcolor{tabG}{0.336} & \textcolor{tabG}{0.384} & \textcolor{tabG}{\underline{0.233}}\\
        & 192 & 0.956 & \textcolor{tabG}{0.394} & \textcolor{tabG}{0.541} & \textcolor{tabG}{0.276} & \textcolor{tabG}{0.269} & \textcolor{tabG}{0.282} & \textcolor{tabG}{0.572} & \textcolor{tabG}{0.544} & \textcolor{tabG}{0.336} & \textcolor{tabG}{0.367} & \textcolor{tabG}{0.544} & \textbf{\textcolor{tabG}{\underline{0.263}}}\\
        & 336 & 1.090 & \textcolor{tabG}{0.452} & \textcolor{tabG}{0.566} & \textcolor{tabG}{0.312} & \textcolor{tabG}{\textbf{0.301}} & \textcolor{tabG}{0.319} & \textcolor{tabG}{0.571} & \textcolor{tabG}{0.566} & \textcolor{tabG}{0.380} & \textcolor{tabG}{0.395} & \textcolor{tabG}{0.523} & \textcolor{tabG}{\underline{0.308}}\\
        & 720 & 0.714 & \textcolor{tabG}{0.479} & \textcolor{tabG}{0.551} & \textcolor{tabG}{0.365} & \textcolor{tabG}{\textbf{0.348}} & \textcolor{tabG}{0.362} & \textcolor{tabG}{0.541} & \textcolor{tabG}{0.559} & \textcolor{tabG}{0.428} & \textcolor{tabG}{0.428} & \textcolor{tabG}{0.741} & \textcolor{tabG}{\underline{0.349}}\\
        \hline
        
    \end{tabular}}%
    \label{tab:dp_test}
\end{table}

In the first set of experiments, we conduct a comparative analysis between the selected Transformer-based and shallow models from the literature and the newly proposed models. The Persistence model is used as a baseline for comparison. The analysis is performed on six datasets, as described in the previous section, and it considers four different forecasting windows.

The results presented in Table~\ref{tab:dp_test} reveal a remarkable finding: the Informer, previously considered to be the best forecasting model up to 2021 and superior to all other Transformer-based models, performs worse than the Persistence model in almost half of the cases, regardless of the forecasting window considered. A similar observation can be made for the Autoformer, although to a lesser extent, and the same holds partially also for the FEDformer. The Crossformer is the only exception within this family of models; it consistently outperforms the non parametric model, as would be desirable, in some cases with a lot of room for improvement.
The Sencoder and Sinformer models show remarkable performance in the forecasting task, consistently outperforming the baseline across all datasets and forecasting windows. In particular, the two proposed models turn out to be the best Transformer-based choice, rivalling the Crossformer model and outscoring it for half of the datasets considered. Despite being significantly simpler than their counterparts, they consistently outperform the state of the art architectures, in some cases by a considerable margin.
When evaluating the performance of the Sencoder and Sinformer models, it is clear that neither model consistently outperforms the other. On average, the Sencoder performs slightly better than the Sinformer, although differences are negligible. As the Sencoder model is the first half of the Sinformer, and therefore has fewer parameters and operations, it should be considered the preferred choice by Occam's razor.

The analysis of the results obtained by shallow models shows that none of them performs worse than the Persistence model across datasets and forecast windows considered. Further analysis of the data presented in Table~\ref{tab:dp_test} shows that the shallow models consistently produce the best overall results. Furthermore, with the exception of the MLP model - which is the most complex of all the shallow models and at the same time the less performing one - no single model emerges as the preferred choice. Rather, they are all interchangeable and equally competitive, consistently producing a very low MAE. This confirms the hypothesis that shallow models are highly effective in generating accurate forecasts, outperforming their Transformer-based counterparts. In addition, it is noteworthy that the MLP model, although not specifically designed for forecasting tasks and often considered unsuitable in the literature, also performs exceptionally well, achieving excellent levels of MAE, particularly for small forecast windows, and competing effectively with any current variant of the Transformer.

\begin{table}[t]
    \caption{This table displays the results of experiments conducted on test sets, calculated as the average percentage improvement per forecasting window between the Mean Absolute Error of the selected models and the Mean Absolute Error of the baseline, represented by the Persistence model. Results that outperform the baseline are coloured \textcolor{tabG}{green}, while those that underperform it are coloured \textcolor{tabR}{red}. The best results are highlighted in \textbf{bold}, while the best results among the Transformer-based models are \underline{underlined}.}
    \centering
    \resizebox{\textwidth}{!}{%
    \begin{tabular}{|c||ccccc|cccccc|}
        \hline
        \raisebox{-6pt}{\textbf{Model}} & \textbf{SLP} & \textbf{MLP} & \textbf{Linear} & \textbf{NLinear} & \textbf{DLinear} & \textbf{Sencoder} & \textbf{Sinformer} & \textbf{FEDformer} & \textbf{Autoformer} & \textbf{Informer} & \textbf{Crossformer}\\
        & (ours) & (ours) & AAAI2023 & AAAI2023 & AAAI2023 & (ours) & (ours) & ICML2022 & NeurIPS2021 & AAAI2021 & ICLR2023\\
        \hline
        96 & \textcolor{tabG}{0.512} & \textcolor{tabG}{0.464} & \textcolor{tabG}{0.511} & \textcolor{tabG}{\textbf{0.518}} & \textcolor{tabG}{0.511} & \textcolor{tabG}{0.460} & \textcolor{tabG}{\underline{0.464}} & \textcolor{tabG}{0.414} & \textcolor{tabG}{0.304} & \textcolor{tabR}{-0.134} & \textcolor{tabG}{0.434}\\
        192 & \textcolor{tabG}{0.494} & \textcolor{tabG}{0.439} & \textcolor{tabG}{0.495} & \textcolor{tabG}{\textbf{0.512}} & \textcolor{tabG}{0.497} & \textcolor{tabG}{\underline{0.439}} & \textcolor{tabG}{0.331} & \textcolor{tabG}{0.419} & \textcolor{tabG}{0.336} & \textcolor{tabR}{-0.223} & \textcolor{tabG}{\underline{0.439}}\\
        336 & \textcolor{tabG}{\textbf{0.403}} & \textcolor{tabG}{0.279} & \textcolor{tabG}{0.363} & \textcolor{tabG}{0.401} & \textcolor{tabG}{0.373} & \textcolor{tabG}{0.308} & \textcolor{tabG}{0.322} & \textcolor{tabG}{0.200} & \textcolor{tabG}{0.065} & \textcolor{tabR}{-0.348} & \textcolor{tabG}{\underline{0.401}}\\
        720 & \textcolor{tabG}{0.331} & \textcolor{tabG}{0.222} & \textcolor{tabG}{0.309} & \textcolor{tabG}{\textbf{0.393}} & \textcolor{tabG}{0.359} & \textcolor{tabG}{0.230} & \textcolor{tabG}{0.212} & \textcolor{tabG}{0.315} & \textcolor{tabR}{-0.006} & \textcolor{tabR}{-0.240} & \textcolor{tabG}{\underline{0.348}}\\
        \hline

    \end{tabular}}%
    \label{tab:average}
\end{table}
\begin{figure}[p]
    \centering
    \label{fig:forecast}
    \resizebox{\textwidth}{!}{%
    \begin{tabular}{ccc}
        & SLP & Sencoder\\
        \raisebox{30pt}{\rotatebox[origin=c]{90}{ETTh1}} &
        \includegraphics[width=7cm]{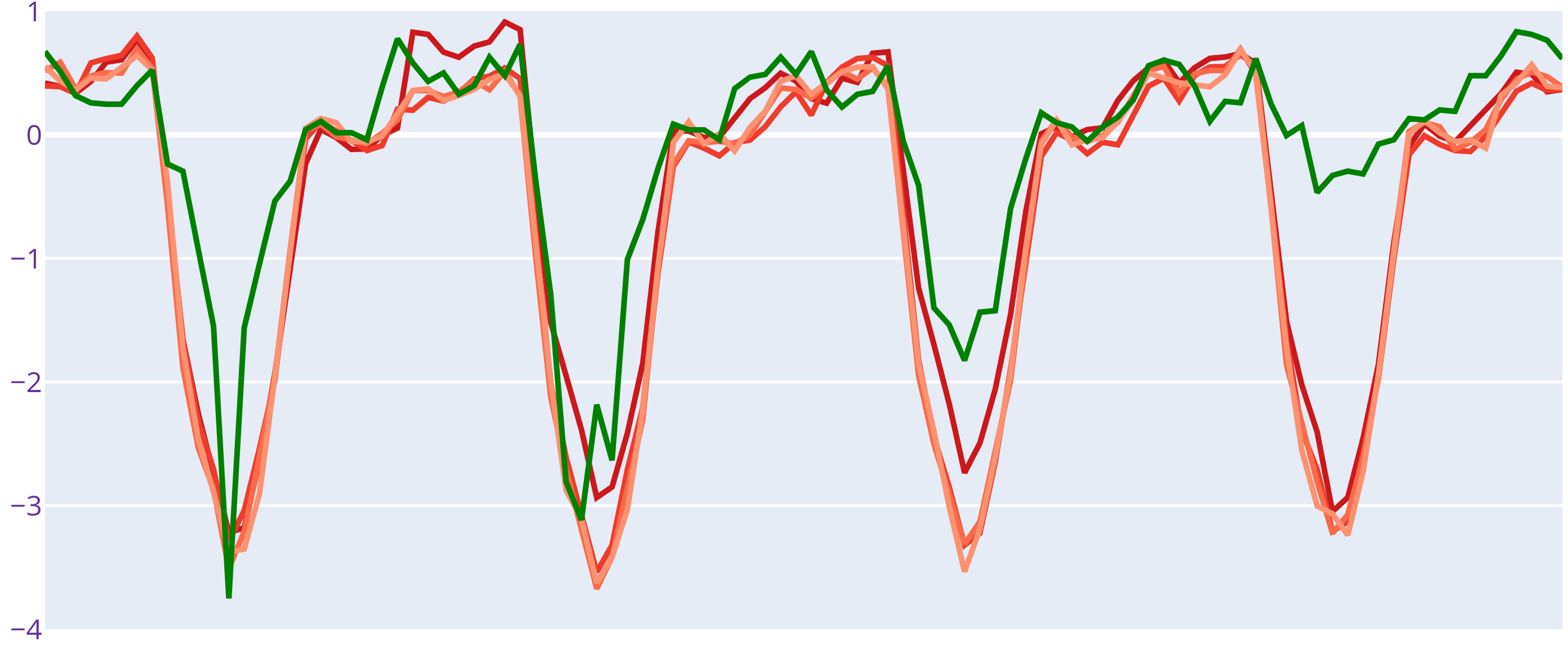}
        &
        \includegraphics[width=7cm]{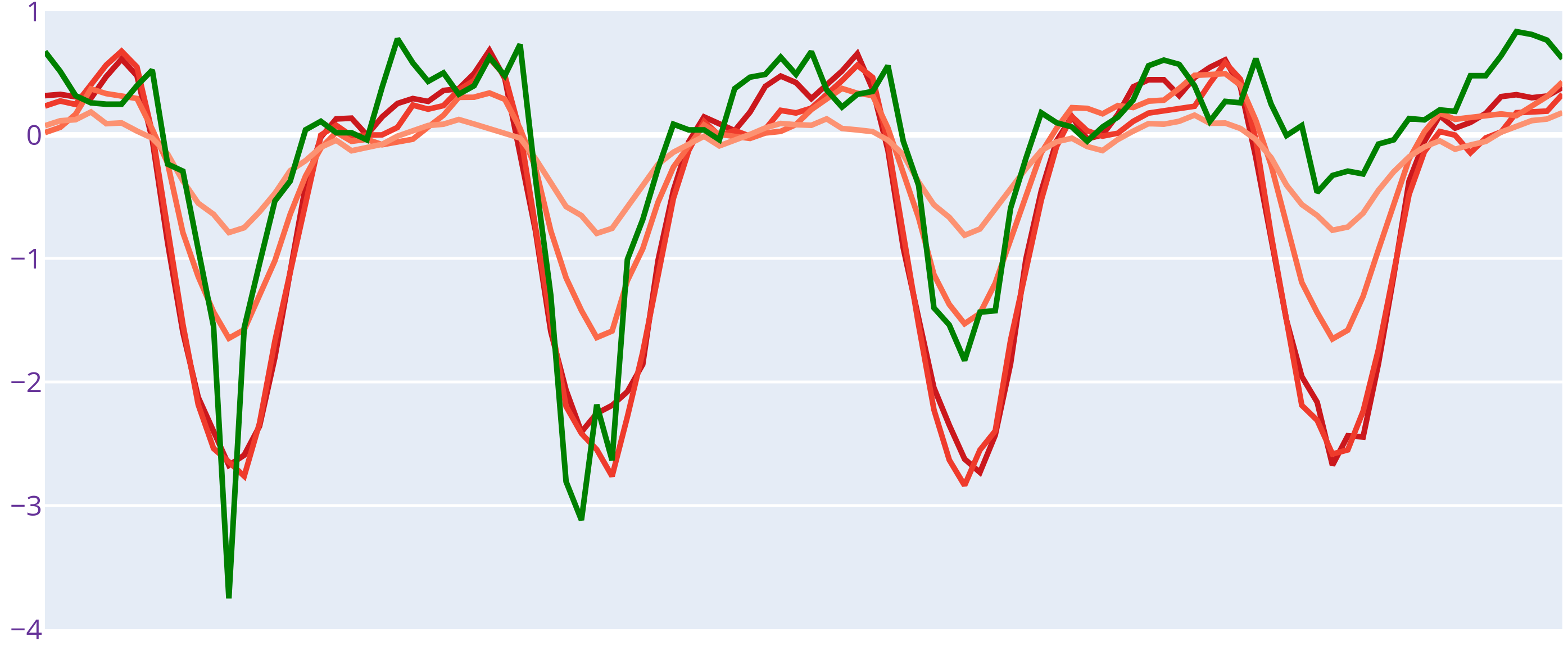}
        \\
        \raisebox{30pt}{\rotatebox[origin=c]{90}{ETTm1}} &
        \includegraphics[width=7cm]{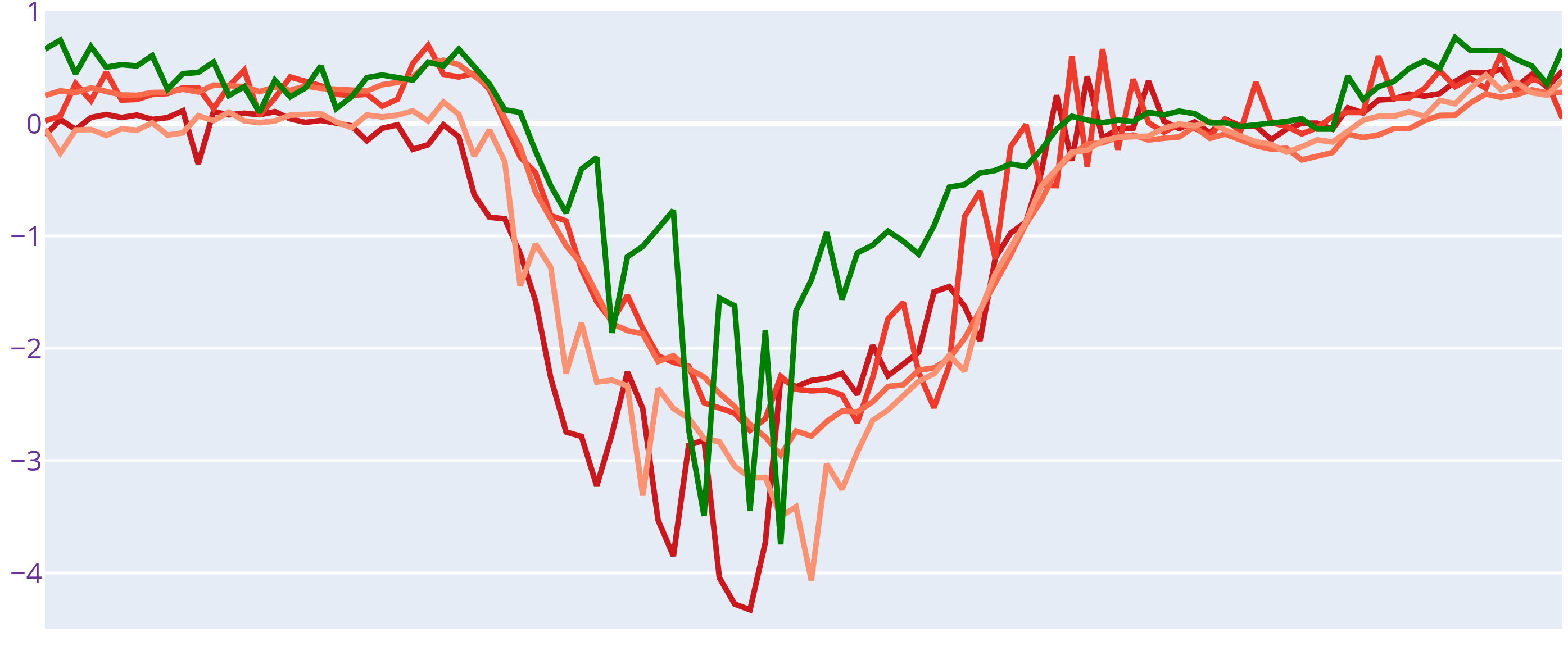}
        &
        \includegraphics[width=7cm]{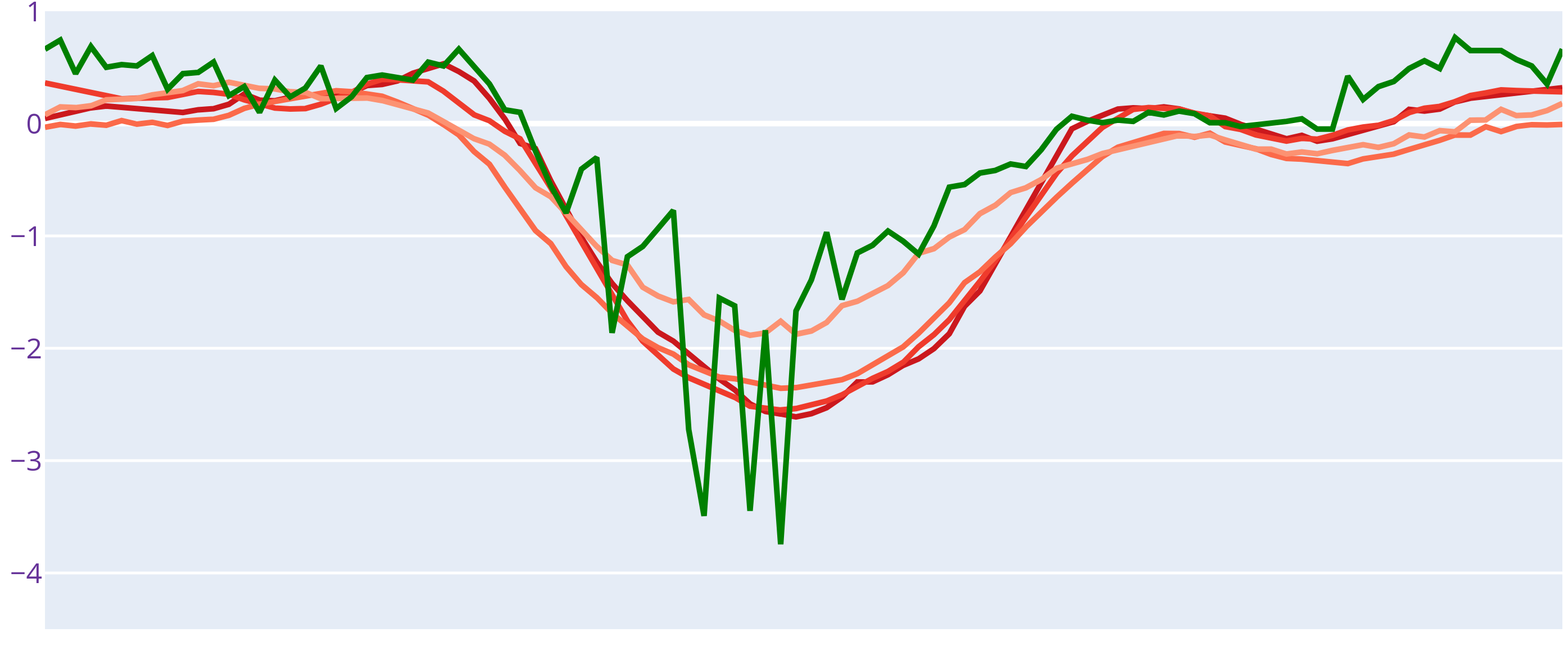}
        \\
        \raisebox{30pt}{\rotatebox[origin=c]{90}{Electricity}} &
        \includegraphics[width=7cm]{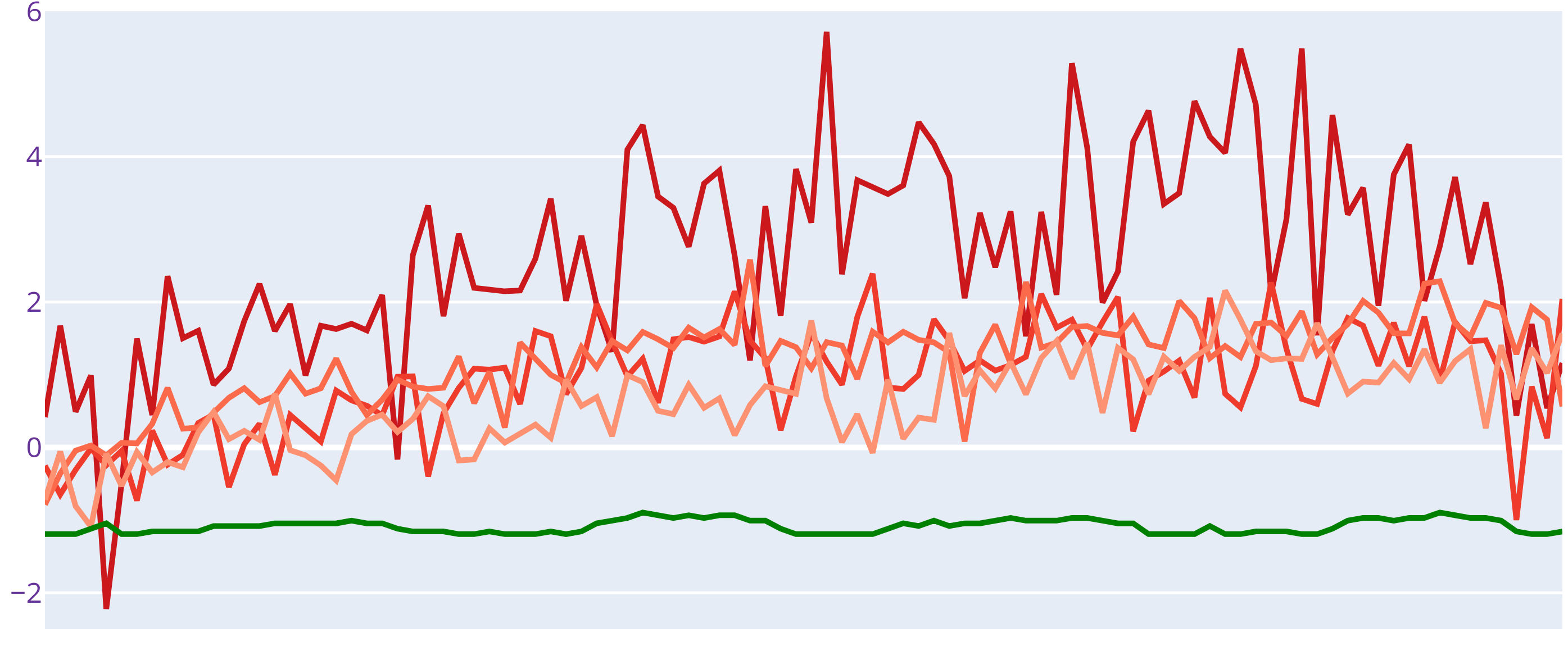}
        &
        \includegraphics[width=7cm]{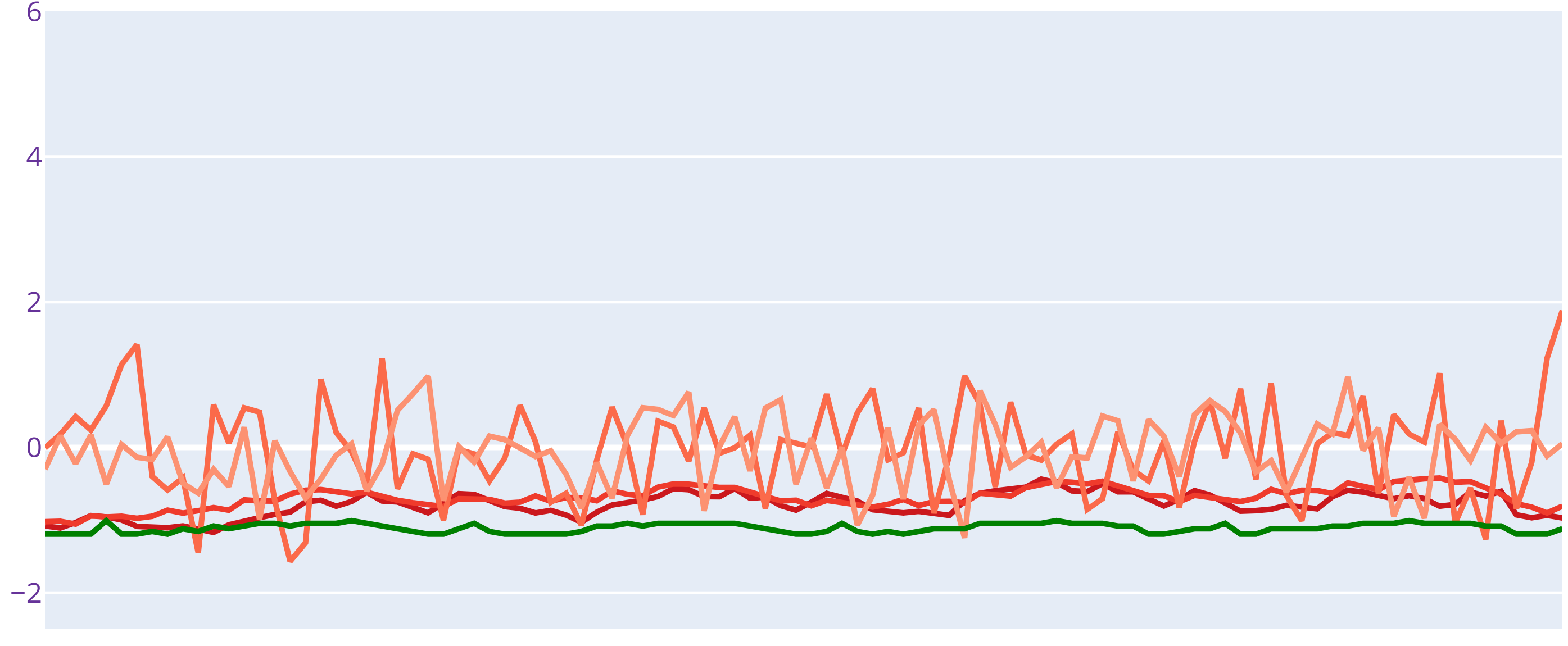}
        \\
        \raisebox{30pt}{\rotatebox[origin=c]{90}{Milan T°}} &
        \includegraphics[width=7cm]{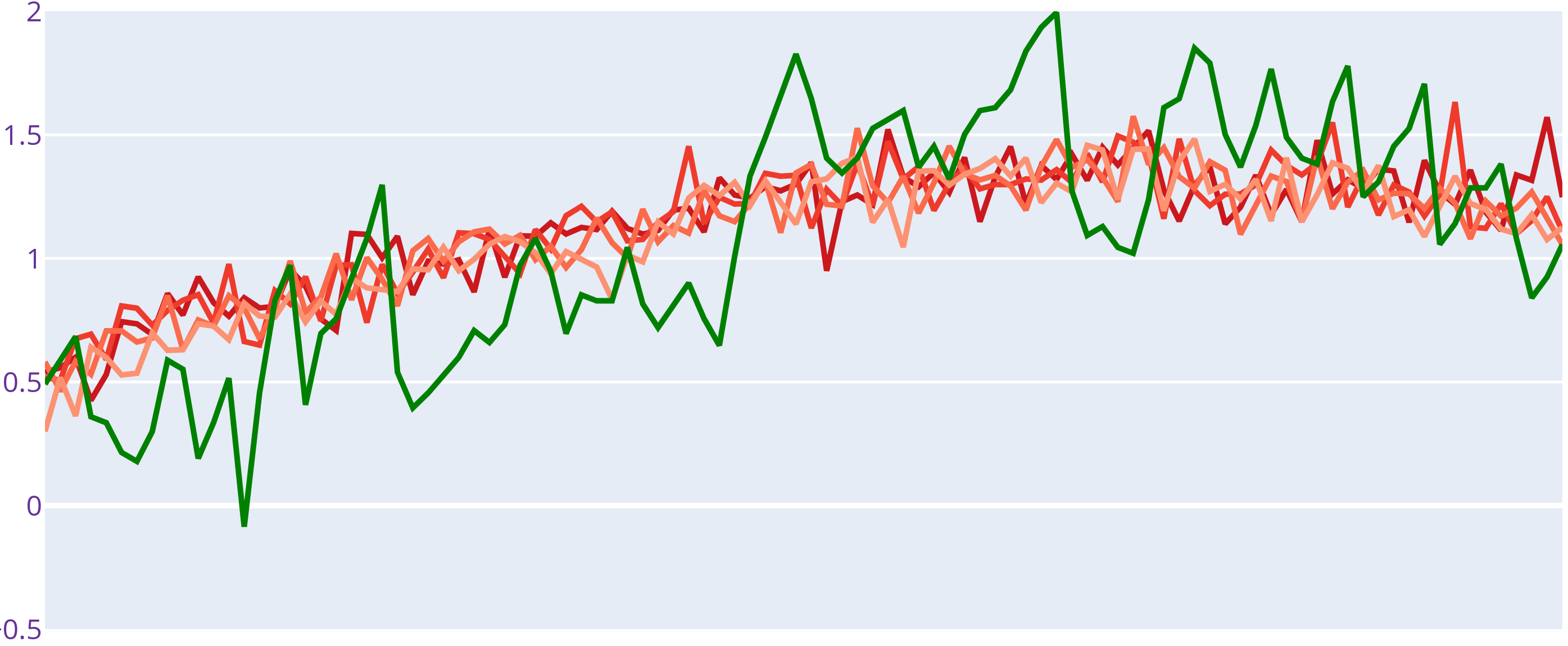}
        &
        \includegraphics[width=7cm]{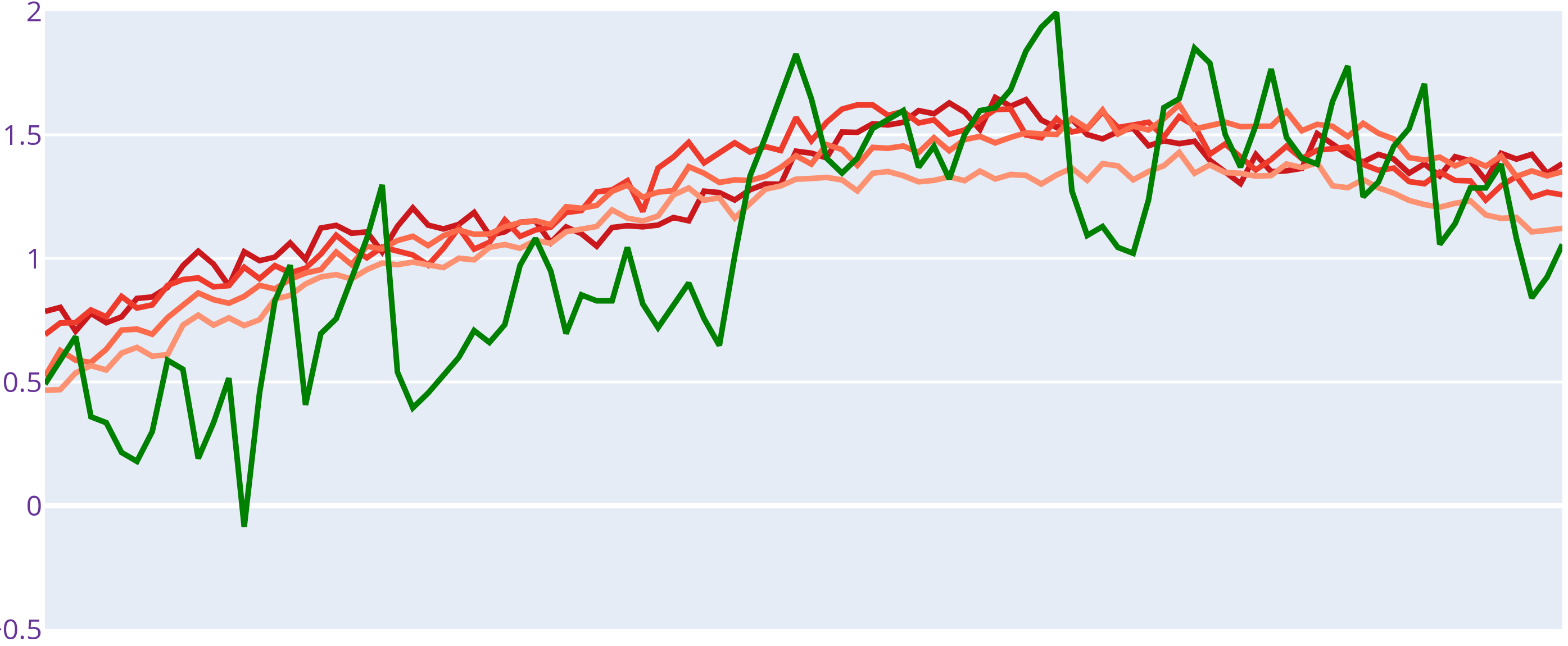}
        \\
        \raisebox{30pt}{\rotatebox[origin=c]{90}{Venice}} &
        \includegraphics[width=7cm]{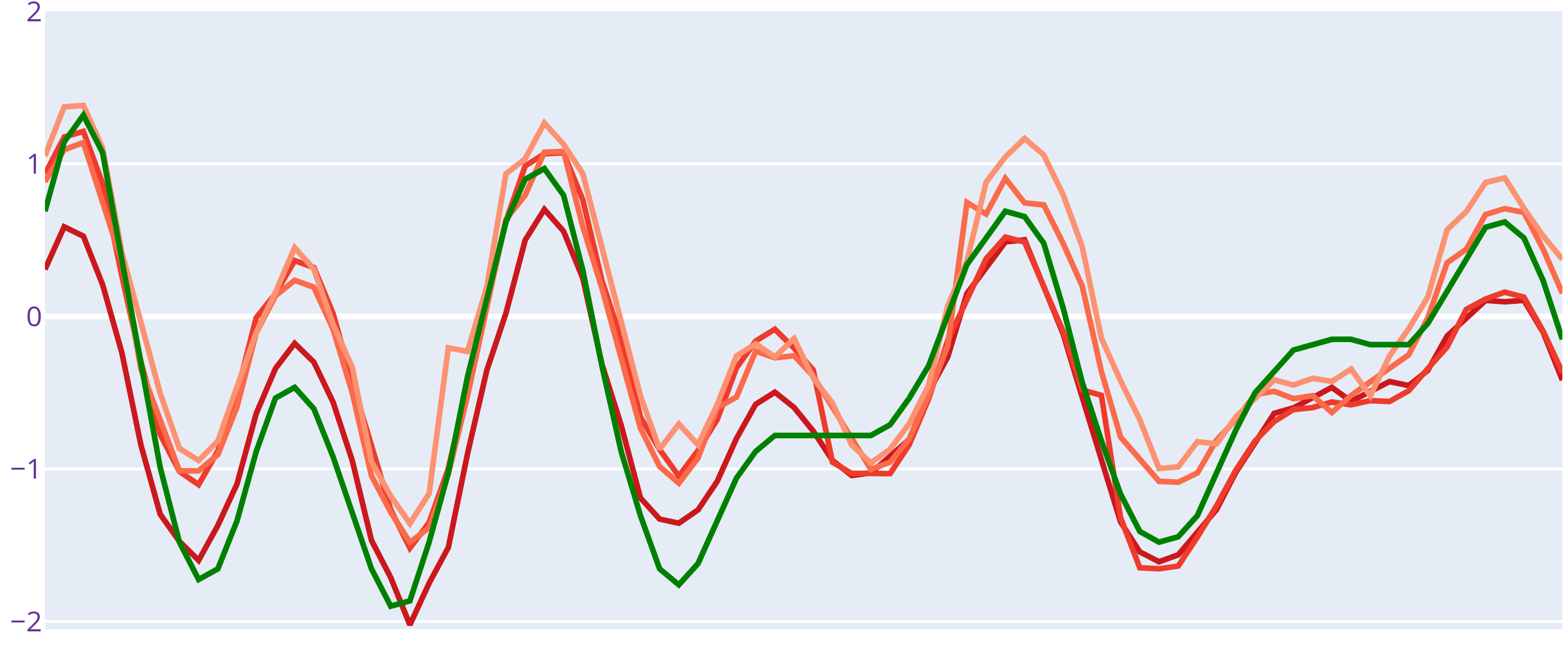}
        &
        \includegraphics[width=7cm]{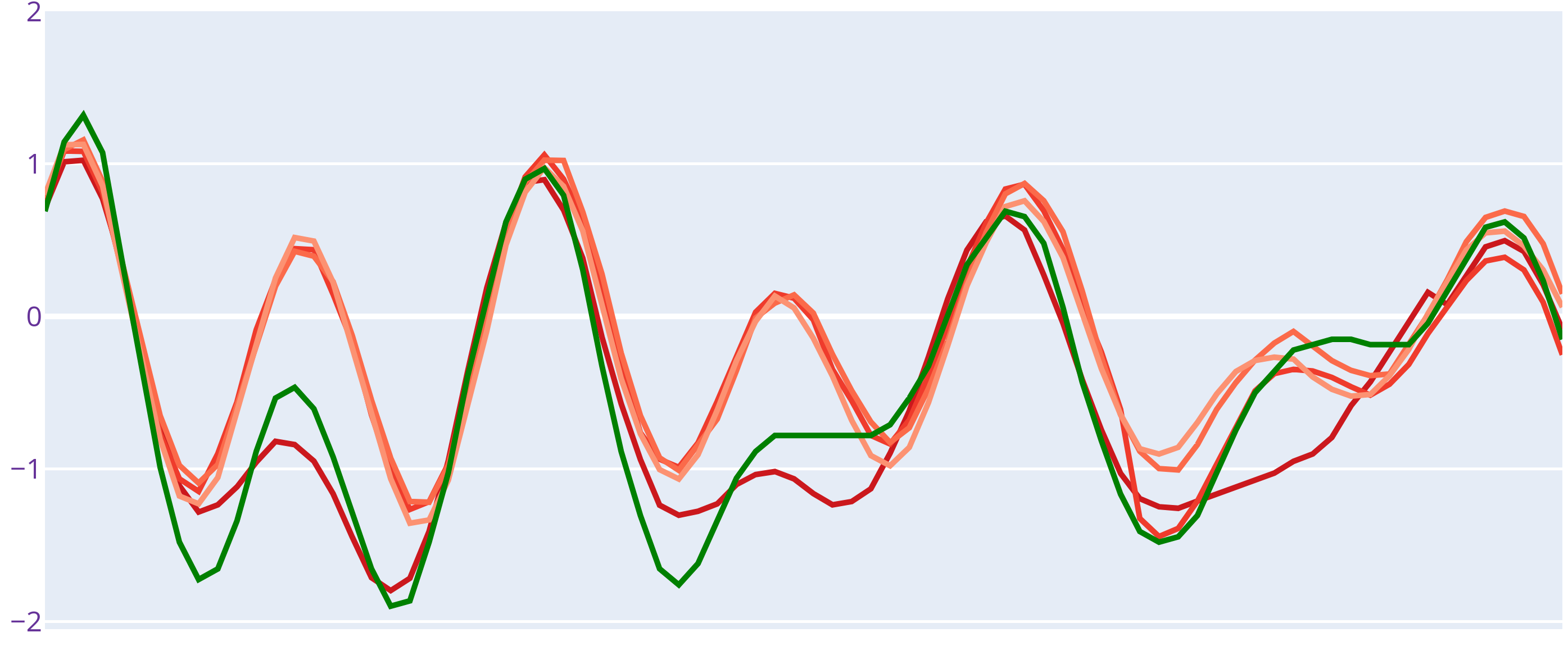}
        \\        
        \raisebox{30pt}{\rotatebox[origin=c]{90}{Weather}} &
        \includegraphics[width=7cm]{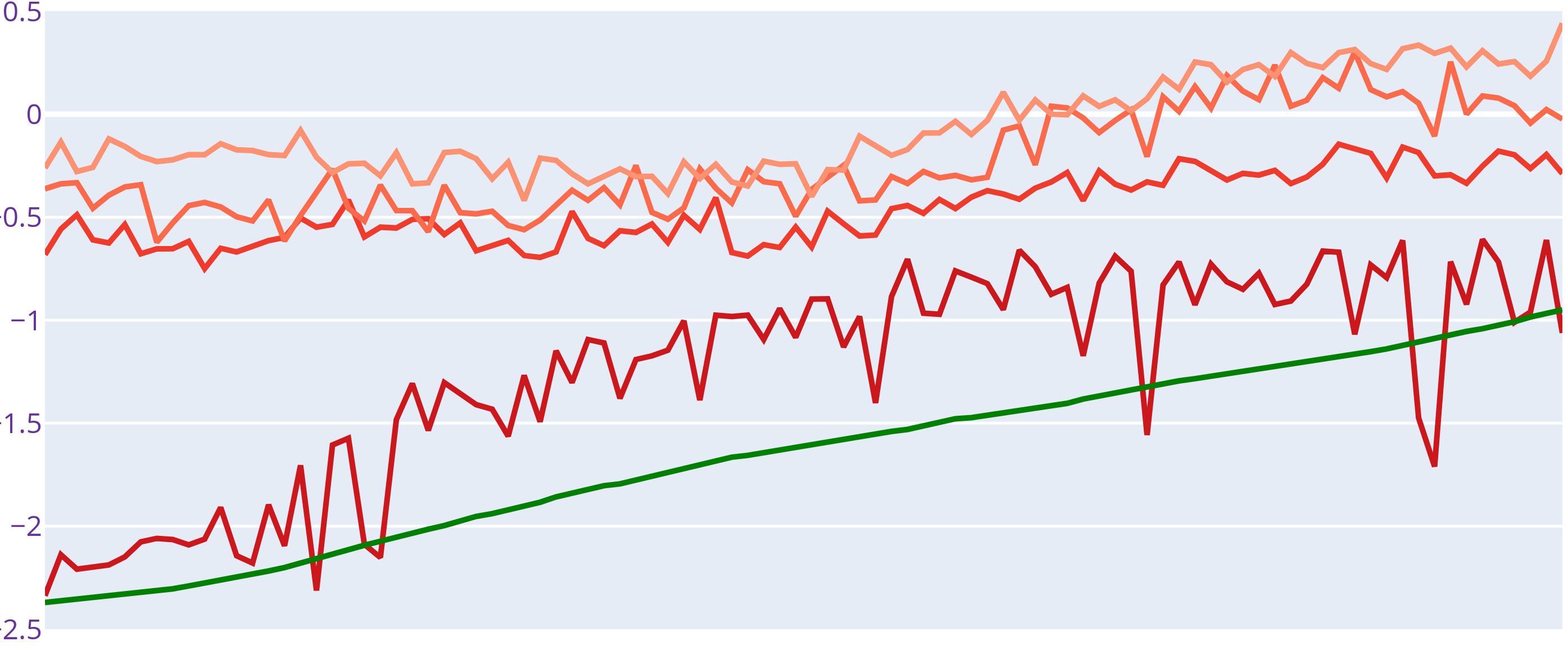}
        &
        \includegraphics[width=7cm]{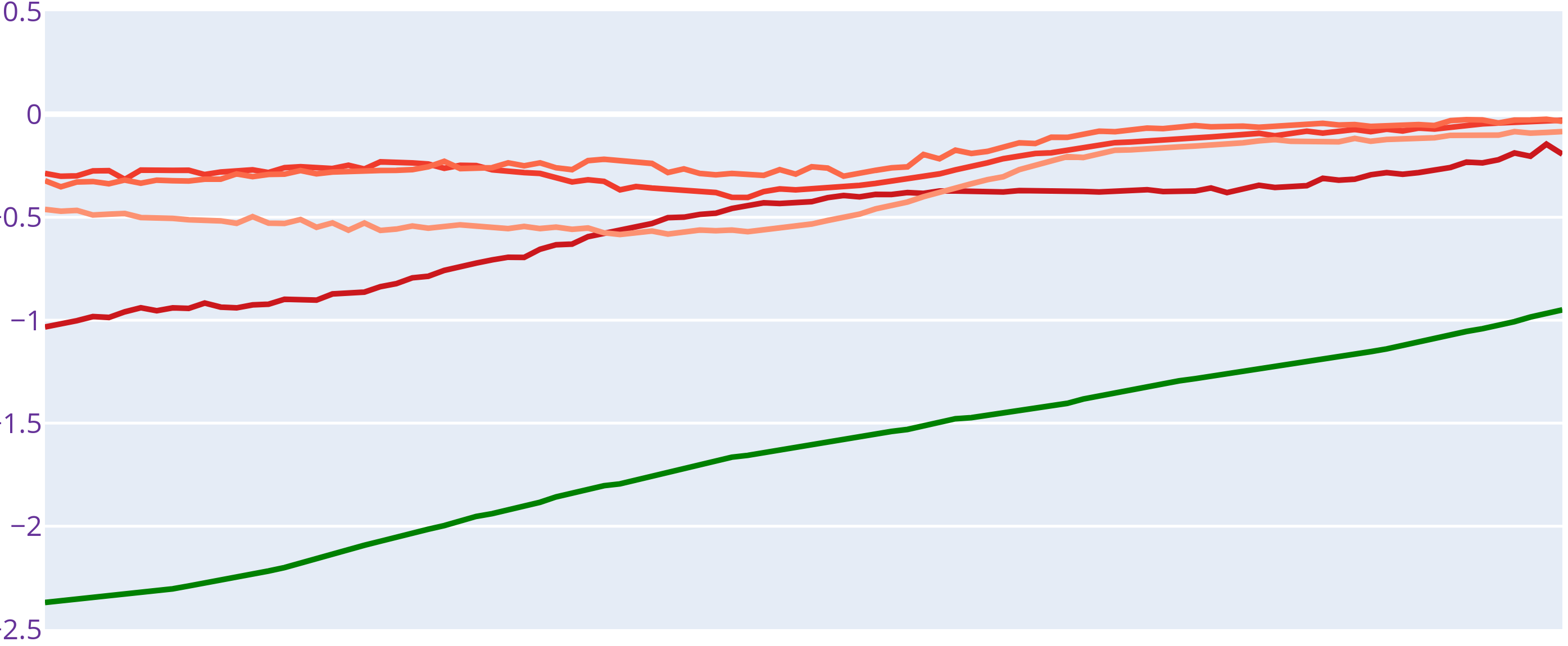}\\
        & & \includegraphics[width=7cm]{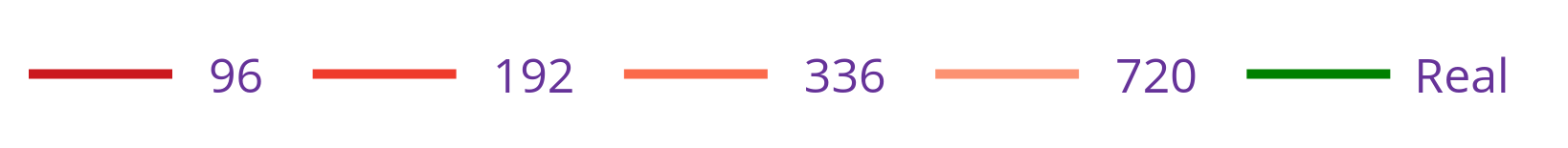}
    \end{tabular}}%
    \caption{This figure depicts a qualitative comparison between the forecasts generated by the SLP model and those produced by the Sencoder model. All predictions were calculated with respect to the datasets test sets and were made across four distinct forecasting windows.}
\end{figure}

To gain a comprehensive understanding of the performance of the models with respect to the size of the forecasting window and to assess their advantage over the baseline, we calculated the percentage improvement between the MAE of the models considered and that of the Persistence model. We then compute the average for each dataset over the different forecasting windows. The results are summarised in Table~\ref{tab:average}.
A first look at the results shows that, in the context of a time series forecasting problem, it is generally preferable to use the Persistence model over the Informer, regardless of the forecasting window considered. Furthermore, it is evident that the Sencoder and Sinformer models are the best Transformer-based options for medium and small windows, reaching the state of the art in this model family, while the Crossformer is preferable for longer forecasts. It is also fair to say that SLP and NLinear are the optimal choices due to their simplicity and superior performance. Overall, the results convey a clear message: on average, any shallow model is preferable to any Transformer-based model.

Figure~1 shows a graphical comparison between SLP and Sencoder on a subset of the test set for each dataset analysed. Looking at all the plots, it is clear that neither SLP nor Sencoder emerges as the clear winner in every context. SLP produces predictions that are accurate but noisy, while Sencoder produces smoother ones. Both models are successful in learning patterns and periodicity, but struggle to make good predictions when these features are less obvious or absent. For the ETTh1, ETTh2 and Venice datasets, the predictions are close to the ground truth in many cases. However, for the Electricity, Milan T° and Weather datasets, the models are less successful in approximating non-periodicity, but still accurately capture the trend and quota of the time series.

Thanks to the availability of a particularly large dataset such as Venice, the last set of experiments concerns the analysis of forecasting when the prediction window becomes extremely long.
Table~\ref{tab:long_forecasting} shows the results of this study.
The first relevant result concerns the absence of the Transformer-based models which, on average, saturate the available memory and thus became intractable already with a window between 1440 and 2880 points and were therefore omitted.
This is certainly another feature to be taken into account; despite much work has been done to optimise the spatial complexity of Transformer-based models, it is still unthinkable to use them for extremely long forecasts.
As far as the shallow models are concerned, many of the considerations made for windows shorter than 720 samples still apply, such as the fact that each of them manages to achieve a smaller error than Persistence.
Although other shallow models achieve good performance, the SLP model proved to be the best among the alternatives analysed. This suggests the effectiveness of the AddT2V embedding and sinusoidal activations to exploit periodicities during the training phase and achieve to good generalisation on unseen data. 
Another general result, but particularly evident from the SLP model results, is the lack of a clear and linear proportionality between prediction length and prediction MAE.
For example, a 240-fold increase in prediction size from 96 to 23040 points resulted in only an 88.8\% increase in error, as shown in Table \ref{tab:dp_test}. This demonstrates the potential of a simple yet powerful model for forecasting medium, long and extremely long time series.

The final message from the analyses conducted in this study is quite clear: at the current state of research, Transformer-based models, while theoretically superior, are not, on average, the best choice for solving time series forecasting problems.
By this we do not mean that it will not be possible to disprove the following statement as research progresses.
What we are saying is that this type of analysis requires a thoroughness that has been lacking in almost all of the work presented in the literature in recent years, and which has led to probably overestimate the Informer, which, in comparison, is not even able to compete with the Persistence model.
Finally, the results obtained with the simplest models, such as the SLP model, suggest that perhaps the best strategy in the field of time series forecasting is to start with simple techniques and use them as a basis for applying methods and strategies currently used in complex models.

\begin{table}[t]
    \caption{This table presents the results of the experiments carried out on the test sets of the Venice dataset in the form of Mean Absolute Error. Evaluations were performed over five different forecasting windows, distinguished by their increased length relative to the standard windows used in most other papers. Results that outperform the baseline are coloured \textcolor{tabG}{green}, while those that underperform it are coloured \textcolor{tabR}{red}. The best results are highlighted in \textbf{bold}, while the second best results are \underline{underlined}.}
    \centering
    \resizebox{0.8\textwidth}{!}{%
    \begin{tabular}{|c||c|ccccc|}
        \hline
        \raisebox{-6pt}{\textbf{Model}} & \textbf{Persistence} & \textbf{SLP} & \textbf{MLP} & \textbf{Linear} & \textbf{NLinear} & \textbf{DLinear}\\
        & (baseline) & (ours) & (ours) & AAAI2023 & AAAI2023 & AAAI2023\\
        \hline
        % 96 & 0.936 & \textbf{\textcolor{tabG}{0.223}} & \underline{\textcolor{tabG}{0.274}} & \textcolor{tabG}{0.275} & \textcolor{tabG}{0.279} & \underline{\textcolor{tabG}{0.274}}\\
        % 192 & 1.190 & \textbf{\textcolor{tabG}{0.242}} & \textcolor{tabG}{0.364} & \textcolor{tabG}{0.325} & \textcolor{tabG}{0.330} & \underline{\textcolor{tabG}{0.322}}\\
        % 336 & 0.525 & \textbf{\textcolor{tabG}{0.252}} & \textcolor{tabG}{0.394} & \textcolor{tabG}{0.359} & \textcolor{tabG}{0.368} & \underline{\textcolor{tabG}{0.357}}\\
        % 720 & 0.540 & \textbf{\textcolor{tabG}{0.262}} & \textcolor{tabG}{0.413} & \underline{\textcolor{tabG}{0.396}} & \textcolor{tabG}{0.410} & \textcolor{tabG}{0.398}\\
        1440 & 0.680 & \textbf{\textcolor{tabG}{0.395}} & \textcolor{tabG}{\underline{0.402}} & \textcolor{tabG}{0.432} & \textcolor{tabG}{0.447} & \textcolor{tabG}{0.432}\\
        2880 & 0.966 & \textbf{\textcolor{tabG}{0.420}} & \textcolor{tabG}{\underline{0.426}} & \textcolor{tabG}{0.477} & \textcolor{tabG}{0.486} & \textcolor{tabG}{0.483}\\
        5760 & 1.125 & \textbf{\textcolor{tabG}{0.410}} & \textcolor{tabG}{\underline{0.418}} & \textcolor{tabG}{0.480} & \textcolor{tabG}{0.501} & \textcolor{tabG}{0.479}\\
        11520 & 1.364 & \textbf{\textcolor{tabG}{0.407}} & \textcolor{tabG}{\underline{0.418}} & \textcolor{tabG}{0.482} & \textcolor{tabG}{0.488} & \textcolor{tabG}{0.481}\\
        23040 & 1.018 & \textbf{\textcolor{tabG}{0.421}} & \textcolor{tabG}{\underline{0.466}} & \textcolor{tabG}{0.472} & \textcolor{tabG}{0.492} & \textcolor{tabG}{0.472}\\
        \hline
    \end{tabular}}%
    \label{tab:long_forecasting}
\end{table}
\section{Conclusion}\label{Conclusion}

In this article we discussed the effectiveness of applying Transformer-based techniques in the context of time series prediction.
The results of the experiments showed that the key to improving these architectures is simplification, and that currently the best performing models are simplified to the point where they are no longer Transformers, but even shallow neural networks.
We discussed the importance of a baseline and showed how the Persistence model is able to outperform models that have been awarded state of the art techniques in recent years.
Finally, we showed how shallow models are able to make accurate predictions of extremely long time series, which are computationally prohibitive for current Transformer-based models due to their polynomial complexity.

Nevertheless, the main objective of this research was not to present new forecasting models, but rather to share with the reader a reflection supported by experimental results.
It seems that in recent years, at least in the time series forecasting field, research has focused more on the desire to apply a particular fashionable technique at all costs than on finding a solution to the problem to be solved. 
Having shown that very simple or even non parametric models are able to outperform algorithms with millions of parameters, is it fair to ask whether part of the scientific community has reached a dead end and talks about it as if it were a highway? We leave it to the reader to answer this question.
Our hope is that, having shown that there are better techniques from a performance point of view, and much better ones from a performance-complexity perspective, it will be possible to continue down the path of finding better solutions by looking forward while keeping an eye on the past. We hope that in this area, as in others, we can move on by taking two steps forward and one behind.

\bibliographystyle{splncs04}
\bibliography{biblio}

\end{document}